\documentclass{article}
\usepackage{ijcai21}

\usepackage{times}

\usepackage{soul}
\usepackage{url}
\usepackage[hidelinks]{hyperref}
\usepackage[utf8]{inputenc}
\usepackage[small]{caption}
\usepackage{graphicx}
\usepackage{amsmath}
\usepackage{booktabs}
\usepackage{amssymb}
\usepackage{colortbl}
\usepackage{multirow}
\usepackage{graphicx}
\usepackage{amsmath,amsthm}
\theoremstyle{definition}
\newtheorem{definition}{Definition}
\newtheorem{problem}{Problem}
\usepackage{xspace,color}
\usepackage{diagbox,multirow} 
\usepackage{subfigure}
\usepackage{tikz,bm}
\usepackage{algorithmic,algorithm}
\definecolor{PineGreen}{rgb}{0.0, 0.47, 0.44}
\definecolor{Gray}{gray}{0.85}
\usepackage{enumitem}
\setlist[itemize]{leftmargin=*}
\usepackage{amsfonts}

\usepackage[utf8]{inputenc}
\usepackage[T1]{fontenc}    % use 8-bit T1 fonts
\usepackage{url}            % simple URL typesetting
\usepackage{booktabs}       % professional-quality tables 
\usepackage{nicefrac}       % compact symbols for 1/2, etc.
\usepackage{microtype}      % microtypography
\usepackage{tikz,bm}
\usepackage{adjustbox}      
\usepackage{filecontents}
\usepackage{array,rotating}
\usepackage[labelfont=bf]{caption}
\newcommand{\RB}{\mathbb{R}}

\newcommand{\bfe}{\mathbf{e}}

\newcommand{\bfx}{\mathbf{x}}

\newcommand{\bfy}{\mathbf{y}}

\newcommand{\calA}{\mathcal{A}}
\newcommand{\calD}{\mathcal{D}}
\newcommand{\calE}{\mathcal{E}}

\newcommand{\calL}{\mathcal{L}}
\newcommand{\calM}{\mathcal{M}}

\newcommand{\calT}{\mathcal{T}}

\newcommand{\mbf}[1]{{\boldsymbol{\mathbf{#1}}}}
\renewcommand{\bm}{\mbf}

\newcommand{\mname}{\texttt{HINT}\xspace}

\newcommand{\stdfontsize}{\tiny}

\title{Uncertainty Quantification on Clinical Trial Outcome Prediction}

\author{
Tianyi Chen$^{1,+}$\and
Yingzhou Lu$^{2,+}$\and 
Nan Hao$^3$\and
Yuanyuan Zhang$^4$\and 
Capucine Van Rechem$^1$\and
Jintai Chen$^{5*}$\and 
Tianfan Fu$^1$\footnote{Corresponding Author}
\affiliations
$+$ Equally contributed \\
$^1$Computer Science Department, Rensselaer Polytechnic Institute\\
$^2$School of Medicine, Stanford University \\
$^3$Stony Brook University Hospital \\
$^4$Purdue University \\
$^5$Computer Science Department, University of Illinois Urbana-Champaign 
\emails
tyc1125@outlook.com, 
hao.nan@stonybrook.edu,
lyz66@stanford.edu, 
cvrechem@stanford.edu,
jtchen721@gmail.com, 
fut2@rpi.edu
}

\begin{document}
\maketitle

\begin{abstract}
The importance of uncertainty quantification is increasingly recognized in the diverse field of machine learning. Accurately assessing model prediction uncertainty can help provide deeper understanding and confidence for researchers and practitioners. This is especially critical in medical diagnosis and drug discovery areas, where reliable predictions directly impact research quality and patient health.
In this paper, we proposed incorporating uncertainty quantification into clinical trial outcome predictions. Our main goal is to enhance the model's ability to discern nuanced differences, thereby significantly improving its overall performance.
We have adopted a selective classification approach to fulfill our objective, integrating it seamlessly with the Hierarchical Interaction Network (HINT), which is at the forefront of clinical trial prediction modeling. Selective classification, encompassing a spectrum of methods for uncertainty quantification, empowers the model to withhold decision-making in the face of samples marked by ambiguity or low confidence, thereby amplifying the accuracy of predictions for the instances it chooses to classify. A series of comprehensive experiments demonstrate that incorporating selective classification into clinical trial predictions markedly enhances the model's performance, as evidenced by significant upticks in pivotal metrics such as PR-AUC, F1, ROC-AUC, and overall accuracy.
Specifically, the proposed method achieved 32.37\%, 21.43\%, and 13.27\% relative improvement on PR-AUC over the base model (HINT) in phase I, II, and III trial outcome prediction, respectively. When predicting phase III, our method reaches 0.9022 PR-AUC scores.
These findings illustrate the robustness and prospective utility of this strategy within the area of clinical trial predictions, potentially setting a new benchmark in the field. 
\end{abstract}

\section{Introduction}

% Clinical trial is an essential phase of drug discovery, evaluating patient responses to treatments such as drugs or drug combinations for specific diseases. Clinical trials are usually expensive, time-consuming, and fraught with low success rates. With the advent of machine learning, leveraging these models to predict clinical trial approvals before their commencement presents a compelling advantage. By identifying likely-to-fail trials in advance, resources can be strategically allocated to more promising trials, potentially accelerating drug discovery and enhancing patient benefits.

Conducting clinical trials is an essential step in the process of developing new medications~\cite{wang2022artificial}. In these trials, the reactions of human subjects to potential treatments (such as individual drug molecules or combinations) are evaluated for specific diseases to assess their safety and effectiveness~\cite{vijayananthan2008importance}. As of 2020, the worldwide market for clinical trials was valued at 44.3 billion, with projections to increase to 69.3 billion by 2028~\cite{grand2021clinical,lu2023machine2}. The financial burden of conducting these trials is substantial, often reaching costs of several hundred million dollars~\cite{martin2017much}. Moreover, these trials typically span several years, partly due to the meticulous and phased approach. Nevertheless, even with this significant investment of resources and time, the success probability of these trials is still relatively low~\cite{peto1978clinical,ledford20114}. Clinical trials can be compromised by various issues, including the drug's ineffectiveness, safety concerns, and flawed clinical trial design~\cite{friedman2015fundamentals}. There has been a surge of studies focusing on how to design better clinical trial mechanisms to enhance clinical trial outcome prediction, among which the Hierarchical Interaction Network (HINT)~\cite{fu2022hint,fu2023automated} stands out as a notable advancement. HINT has greatly enhanced the probability of clinical trial outcome prediction before the trial commences, allowing more resources to be allocated to trials that are more likely to succeed by avoiding inevitable failures. However, even in the face of these advancements, trials may still be predicted even if the confidence is not high for them in some uncertain cases. Fortuitously, historical literature suggests that certain algorithms for uncertainty quantification have opened new opportunities in this field. Meanwhile, the extensive historical data on clinical trials and the comprehensive databases on both successful and unsuccessful drugs pave the way for employing machine learning models. This raises a pivotal question: could we utilize the online database and adopt different strategies based on the degree of certainty, thereby increasing the overall clinical trial outcome prediction probability?

Despite the HINT model being the current state-of-the-art method in clinical trial outcome prediction, eclipsing other methodologies in several aspects, there remains scope for enhancement, particularly in terms of accuracy and false alarm rate. For example, the application of machine learning in the medical field necessitates not only reliance on model predictions but also a critical assessment of the model's confidence and timely human intervention. This underscores the vital necessity of not solely depending on the HINT model, as excessive dependence on machine predictions without adequate checks can lead to significant risks. Such risks highlight the importance of integrating uncertainty quantification into these models. Regarding uncertainty quantification, various approaches exist, including Bayesian methods, ensemble techniques, evidential frameworks, Gaussian processes, and conformal prediction.

Among the various methods, conformal prediction stands out due to its simplicity and generality in creating statistically rigorous uncertainty sets for model predictions. {A key feature of these sets is their validity in a distribution-free context, offering explicit, non-asymptotic guarantees independent of any distributional or modeling assumptions~\cite{nemani2023uncertainty}. } 
Nonetheless, the conventional application of conformal prediction in binary classification scenarios has limitations. {Specifically, the resulting prediction lacks practical value when a model predicts both positive and negative outcomes for a sample due to uncertainty.} To address this, we propose a shift towards selective classification, wherein the model offers predictions only when it has high confidence; otherwise, it abstains from yielding a prediction. This approach can be applied to any pre-trained model, ensuring that the model’s predictions are highly probable and specified by human-defined criteria. This method, however, introduces a trade-off between coverage and accuracy, often characterized by a strong negative correlation. Careful consideration of this balance is crucial in practical applications, especially in the sensitive context of medical predictions.

{In this paper, we enhance the Hierarchical Interaction Network (HINT) for general clinical trial outcome prediction tasks. We point out the limitation of the current HINT model and adopt the method of selective classification to quantify the model uncertainty and improve its prediction performance. Empirical experiments indicate that by applying selective classification to HINT, the model demonstrates significant elevations in key metrics on phase-level outcome prediction. 
This work paves the way for future explorations into more nuanced models that can handle the complexities of clinical trial data, offering a beacon for forthcoming research in the field. Our findings advocate for the continued development and refinement of models like HINT, emphasizing the need for precision and care in predictive analytics within clinical research. The potential for these advancements can significantly impact patient outcomes and improve the efficiency of trial design. }

% {A machine learning-based clinical trial outcome prediction model could help improve clinical trials in three different ways: 
% \begin{enumerate}
% \item  Cost Estimation: An integral part of our model is its ability to provide more precise cost estimations for clinical trials. By analyzing factors such as trial duration, the number of participants, and resource needs, the model forecasts the financial requirements with good accuracy. This can aid in better budget allocation and financial planning.
% \item Risk Minimization: Our model significantly contributes to identifying and minimizing potential risks in clinical trials. By leveraging comprehensive analysis, the model identifies possible challenges like drug ineffectiveness, safety concerns, or protocol design flaws early in the trial process. 
% \item Trials Selection Optimization: By assessing various factors, the model guides toward clinical trials that are more likely to produce positive outcomes, thus optimizing resource utilization and accelerating drug development. 
% \end{enumerate}} 

\section{Methods}

\subsection{Formulation}
\label{sec:formulation}
%%% rephrase
% A \underline{\textit{clinical trial}} is designed to validate the safety and efficacy of a \underline{\it treatment set}  towards a \underline{\it target disease set} on a patient group defined by the \underline{\it trial protocol}.
A \underline{\textit{clinical trial}} is an organized research effort to evaluate the safety and efficacy of a \underline{\it treatment set} aimed at combating a \underline{\it target disease set}. This is all guided by a detailed plan known as \underline{\it trial protocol}, which is applied to a select group of patients. The trial aims to understand how the treatment performs across various patients, assessing not only its effectiveness but also identifying any potential side effects. Now we formulate them into more detailed terms for better understanding.

\begin{definition}[\textbf{Treatment Set}]
Imagine our trial has $K_{\tau}$ drug candidates. These form our Treatment Set, denoted as \underline{$\mathcal{T} = \{ \tau_1, \cdots, \tau_{K_{\tau}} \}$}, where $\tau_1, \cdots, \tau_{K_{\tau}}$ are $K_{\tau}$ drug molecules being tested in this trial. Our study concentrates on trials to identify new applications for these drug candidates, while trials focusing on non-drug interventions like surgery or device applications are considered outside the scope of this research.
\begin{equation}
\label{eqn:drug}
\mathcal{T} = \{\tau_1, \cdots, \tau_{K_{\tau}}\},
\end{equation}
\end{definition}

% \begin{definition}[\textbf{Treatment Set}]
% Treatment set includes one or multiple drug candidates, denoted by \underline{$\mathbb{M} = \{{m}_1, \cdots, {m}_{N_m}\}$},
% % \begin{equation}
% % \label{eqn:molecule}
% % \mathbb{M} = \{{m}_1, \cdots, {m}_{N_m}\},
% % \end{equation}
% % \end{definition}
% where $m_1, \cdots, m_{N_m}$ are $N_m$ drug molecules involved in this trial. Our attention is centered on clinical trials with the objective of discovering new indications for drug candidates. Trials that do not incorporate drugs, including surgical procedures and medical devices, fall beyond the purview of our current study and may be addressed in subsequent research works.
% %%% rephrase
% % Note that we focus on clinical trials that aim at discovering new indications of drug candidates. Other trials that do not involve drugs, such as surgeries and devices, are out of scope and can be considered future work.
% \end{definition}

\begin{definition}[\textbf{Target Disease Set}]
\label{def:disease}
This refers to the specific diseases the trial is targeting. If our trial is addressing $K_{\delta}$ diseases, then our Target Disease Set is represented by \underline{$\mathcal{D} = \{\delta_1, \cdots, \delta_{K_{\delta}}\}$}, with each $\delta_i$ symbolizing the diagnostic code \footnote{In this paper, we use ICD10 codes (International Classification of Diseases)~\cite{anker2016welcome}} for the $i$-th disease.
\begin{equation}
\label{eqn:disease_advanced}
\mathcal{D} = \{\delta_1, \cdots, \delta_{K_{\delta}}\},
\end{equation}
\end{definition}

% \begin{definition}[\textbf{Target Disease Set}]
% \label{def:disease}
% Each trial targets one or more diseases. Suppose there are $N_d\geq 1$ diseases \underline{$\mathbb{D} = \{d_1,\cdots, d_{N_d}\}$} in a trial, 
% we represent the target disease set as 
% \begin{equation}
% \label{eqn:disease}
% \mathbb{D} = \{d_1,\cdots, d_{N_d}\}, 
% \end{equation}
% where $d_i$ represents the diagnosis code\footnote{In this paper, we use ICD10 codes (International Classification of Diseases)~\cite{anker2016welcome}} of the $i$-th disease.
% % $d_i$ represents the raw information associated with the disease, including the disease name, description (text data) and its corresponding diagnosis code (e.g., International Classification of Diseases - ICD codes~\cite{anker2016welcome}). 
% \end{definition}

\begin{definition}[\textbf{Trial Protocol}]
The Trial Protocols are the guideline plan of the clinical trial. They are articulated in unstructured natural language and encompass both inclusion (+) and exclusion (-) criteria, which respectively outline the desired and undesirable attributes of potential participants. Imagine a clinical trial for a new medication to treat high blood pressure. Inclusion criteria (+) are the requirements participants need to meet to join the trial. {This might include:
``Adults aged 30-65 years'' or ``Diagnosed with high blood pressure''. Similarly, Exclusion criteria (-) are the factors that would disqualify someone from participating. This might include: ``Pregnant or breastfeeding women'' or ``Those undergoing treatment for cancer.''} By this way, these criteria provide details on various key parameters such as age, gender, medical background, the status of the target disease, and the present health condition.
\begin{equation}
\label{eqn:protocol_advanced}
\quad \mathcal{P} = [\bm{\pi}_{1}^{+}, ..., \bm{\pi}_{Q}^{+}, \bm{\pi}_{1}^{-}, ..., \bm{\pi}_{R}^{-}],\ \ \ \ \ \ \bm{\pi}_{k}^{+/-} \ \text{is a criterion}.
\end{equation}
 $Q$ ($R$) is the number of inclusion (exclusion) criteria in the trial. The term $\bm{\pi}_{k}^{+}$ ($\bm{\pi}_{k}^{-}$) designates the $k$-th inclusion (exclusion) criterion within the trial protocol. Each criterion $\bm{\pi}$ is a sentence in unstructured natural language. 
\end{definition}

{
\begin{definition}[Clinical Trial Approval]
Clinical trial approval refers to a drug passing a certain phase of a clinical trial, which means that the drug has met specific predefined objectives or endpoints for that phase, demonstrating its safety, efficacy, tolerability, or a combination thereof, depending on the trial's goals. Each phase of clinical trials has distinct purposes and criteria for success. 
\end{definition}}

\begin{problem}[\textbf{Clinical Trial Outcome Prediction}]
Predicting whether a clinical trial will get approval is like forecasting the outcome of a complex process. {Clinical trial approval is represented as {a binary label}, where 1 means the trial was a success and 0 means it was not.} The approval of a clinical trial is represented as a binary label \( \omega \in \{0,1\} \), where \( \omega =1 \) signifies a successful trial, and \( 0 \) a failed one. The estimation of \( \omega \), represented as \( \hat{\omega} \), can be formulated through the function \( h_{\xi} \), such that \( \hat{\omega} = h_{\xi}(\mathcal{T}, \mathcal{D}, \mathcal{P}) \), where \( \hat{\omega} \in [0,1] \) denotes the calculated probability of a successful approval. In this context, \( \mathcal{T} \), \( \mathcal{D} \), and \( \mathcal{P} \) refer to the treatment set, the target disease set, and the trial protocol, respectively.
\end{problem}

% \begin{problem}[\textbf{Trial outcome prediction}]
% The approval of the trial is represented by a binary label \( y \in \{0,1\} \), where \( y=1 \) indicates a successful trial, and \( 0 \) indicates failure. The estimation of \( y \), denoted as \( \hat{y} \), can be modeled with the function \( f_{\theta} \), such that \( \hat{y} = f_{\theta}(\mathbb{M}, \mathbb{D}, \mathbb{C}) \), where \( \hat{y} \in [0,1] \) represents the predicted probability of success. Here, \( \mathbb{M} \), \( \mathbb{D} \), and \( \mathbb{C} \) represent the treatment set, the target disease set, and the trial protocol, respectively.
% %%% rephrase
% % The trial approval is a binary label $y\in\{0,1\}$, $y=1$ indicates trial success, and 0 indicates failure. 
% % The estimation of $y$, denoted as $\haty$, can be learned using $f_{\theta}: \haty = f_{\theta}(\mathbb{M}, \mathbb{D}, \mathbb{C})$,
% % where $\haty \in [0,1]$ is predicted success probability, $\mathbb{M}, \mathbb{D}, \mathbb{C}$ are the treatment set, target disease set and trial protocol, respectively.
% \end{problem}

\paragraph{Significance of Clinical Trial Outcome Prediction}

The clinical trial stands out as the most time-consuming and expensive stage in the drug discovery process. Leveraging machine learning for trial optimization and design holds the potential to significantly accelerate the delivery of life-saving therapeutics to patients. {Machine learning tools can play a crucial role in proactively notifying practitioners of potential trial challenges, identifying risks, optimizing safety monitoring protocols, and ensuring participant well-being.} Additionally, these tools can aid in pinpointing suitable patient populations, optimizing sample sizes, refining inclusion and exclusion criteria, and selecting appropriate endpoints and outcome measures.

\subsection{Base Model: \mname}
\label{sec:overview} 

This section describes HINT~\cite{fu2022hint} as the base model. 
Depicted in Figure~1, \mname{} stands as an end-to-end framework, which is innovatively designed to predict the probability of success for a clinical trial before its commencement~\cite{fu2022hint,fu2023automated}.
In the first instance, \mname{} integrates an \textbf{input embedding module}, where it adeptly encodes multi-modal data from various sources, encompassing drug molecules, detailed disease information, and trial protocols into refined input embeddings (Section~\ref{sec:input}).
Thereafter, these embeddings are fed into the \textbf{knowledge embedding module} to synthesize knowledge embeddings that are pretrained using external knowledge (Section~\ref{sec:knowledge}).
Lastly, the \textbf{interaction graph module} serves as a nexus, binding these embeddings through an extensive domain knowledge network. This comprehensive interlinking not only unravels the complexity inherent in various trial components but also maps their multifarious interactions and their collective impact on trial approvals.
Utilizing this foundation, \mname learns a \textit{dynamic attentive graph neural network} to prognosticate the trial approval (Section~\ref{sec:gnn}).

%%% rephrase
% \mname is an end-to-end framework for predicting the success probability of a trial before the trial starts. 
% First, \mname has an {\bf input embedding module} to encode multi-modal data from various sources including drug molecules, disease information and trial protocols to input embeddings (Section~\ref{sec:input}). Next, these embeddings will be fed into the {\bf knowledge embedding module} to generate knowledge embeddings that are pretrained using external knowledge (Section~\ref{sec:knowledge}). Then the {\bf interaction graph module} will connect all the embeddings via domain knowledge to fully capture various trial components and their complex relations as well as their influences on trial approvals. Based on that, \mname learn a \textit{dynamic attentive graph neural network} to predict trial approval (Section~\ref{sec:gnn}).
% An {\bf imputation module} is designed to handle missing data (Section~\ref{sec:imputation}).
% For ease of exposition, we list mathematical notations in Table~\ref{table:notation}.

\begin{figure*}[h!]
\centering
\includegraphics[width=\textwidth]{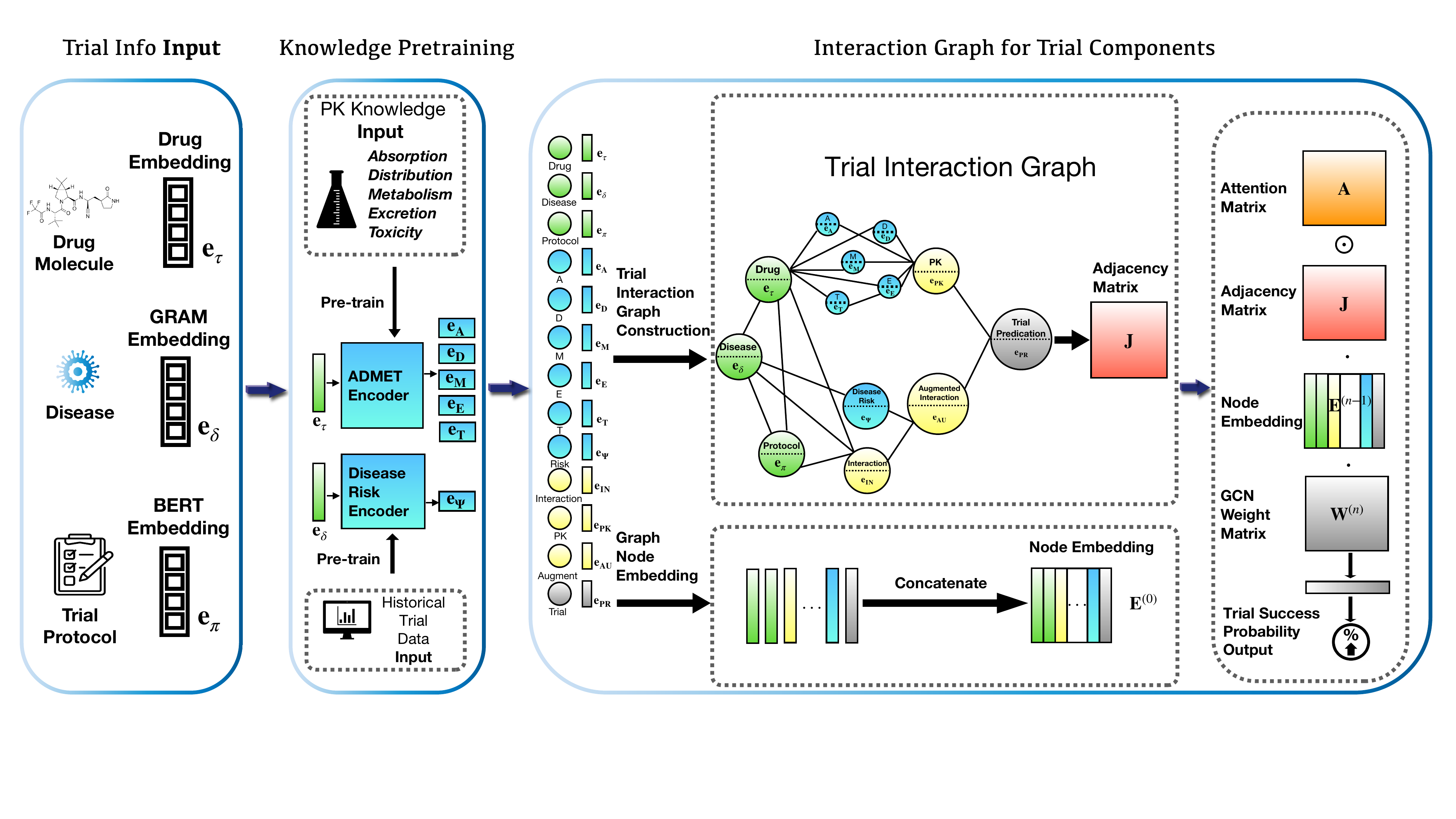}
\vskip -4 em
\caption{\mname extracts features from the following trial components: drug molecule embedding $\bfe_\tau$, disease embedding $\bfe_\delta$, and trial protocol embedding $\bfe_\pi$ (as described in section~\ref{sec:input}). Before constructing an interaction graph using these components, \mname pretrains certain embeddings (depicted as blue nodes) using external knowledge about medication characteristics and disease risks(Section~\ref{sec:knowledge}). Subsequently, we create an interaction graph in Section~\ref{sec:gnn} to depict the interactions among different trial components. Using this interaction graph, we obtain trial embeddings that represent the trial components and their interactions. Leveraging the learned embeddings, we make predictions for trial approvals.}
\label{fig:method}
\end{figure*}

% \subsection{(I) Trial Representation Learning based on Interaction Graph of Trial Components}
\subsection{Input Embedding Module}
\label{sec:input}

{The input of the model contains three data sources: (1) drug molecules, (2) disease information, and (3) trial protocols.} 

\smallskip
\noindent{\bf Drug molecules} play a crucial role in forecasting the approvals of clinical trials. These molecules are typically represented through SMILES strings or molecular graphs~\cite{zhang2021ddn2,wu2022cosbin}.
% There are many existing works in embedding drug molecules into latent vectors including knowledge base approach such as Morgan fingerprint and its variants~\cite{cereto2015molecular}, or representation learning methods for SMILES strings~\cite{wang2019smiles} and molecule graphs~\cite{dai2016discriminative,coley2017convolutional,hu2019strategies}.
% \mname supports all these types of molecule embeddings. 
Formally, treatment set {$\mathcal{T} = \{ \tau_1, \cdots, \tau_{K_{\tau}} \}$} is represented as 
% \begin{equation}
% \label{eqn:molecule_feature}
% \text{\underline{Drug Embedding}}\quad \bfh_m = \frac{1}{N_m}\sum_{j=1}^{N_m} f_m(m_j) \in \RB^{d}, 
% \end{equation}
\begin{equation}
\label{eqn:molecule_feature}
\text{\underline{Drug Embedding}}\quad \bfe_{\tau} = \frac{1}{K_{\tau}}\sum_{i=1}^{K_{\tau}} g_{\tau}(\tau_i) \in \RB^{d},
\end{equation}
where $g_{\tau}(\cdot)$ is designated as the molecule embedding function. By aggregating the molecular embeddings derived from a trial, we obtain {\it drug embedding vector}, which is conceptualized as the mean of all molecular embeddings~\cite{fu2021mimosa,fu2020core}. Our empirical investigations reveal that employing an averaging method as the aggregation mechanism for drug embeddings yields more effective results than utilizing a summative approach.

\smallskip
\noindent{\bf Disease information} can significantly impact trial approvals. For instance, oncology drugs exhibit lower approval rates compared to those for infectious diseases~\cite{Hay2014-qw,fu2019pearl,fu2021differentiable}. Disease information is primarily sourced from its descriptive texts and corresponding ontology, such as disease hierarchies like the International Classification of Diseases(ICD)~\cite{anker2016welcome}. Target Disease Set {$\mathcal{D} = \{\delta_1, \cdots, \delta_{K_{\delta}}\}$} (Definition~\ref{def:disease}) in the trial can be represented as
% \vspace{-2mm}
% \begin{equation}
% \label{eqn:disease_feature}
% \text{\underline{Disease Embedding}}\quad\ \bfh_d = \frac{1}{N_d}\sum_{i=1}^{N_d} \text{GRAM}(d_i) \in \RB^{d},   
% \end{equation}

\begin{equation}
\label{eqn:disease_feature}
\text{\underline{Disease Embedding}}\quad\ \bfe_{\delta} = \frac{1}{K_{\delta}}\sum_{j=1}^{K_{\delta}} G_{\delta}(\delta_j) \in \RB^{d},
\end{equation}

where $G_{\delta}({\delta}_j)$ represents an embedding of disease ${\delta}_j$ using GRAM (graph-based attention model)~\cite{choi2017gram}, which leverages the hierarchical information inherent to medical ontologies. 

\smallskip
\noindent{\bf Trial protocol} is a key document that outlines the conduct of a clinical trial and encompasses specific eligibility criteria essential for patient recruitment. These inclusion or exclusion criteria are systematically articulated in individual sentences. To effectively represent each sentence within these criteria, we utilize Clinical-BERT~\cite{alsentzer2019publicly}. The derived sentence representations are then sequentially processed through 4 one-dimensional convolutional layers~\cite{you2018end}, each layer employing varying kernel sizes to discern semantic nuances at four distinct levels of granularity. This is followed by a fully-connected layer that culminates in the formation of the protocol embedding. Concisely, the protocol embedding is characterized as
\begin{equation}
\label{eqn:protocol_embedding}
\text{\underline{Protocol Embedding}}\quad\ \bfe_{\pi} = g_{\pi}(\mathcal{P}), \ \ \ \ \bfe_{\pi} \in \RB^{d}. 
\end{equation}

\subsection{Pretraining Using External Knowledge}
\label{sec:knowledge}
\mname integrates external knowledge sources to pretrain knowledge nodes and further refine and augment these input embeddings.

\smallskip
\noindent{\bf Pharmaco-kinetics Knowledge:} We engage in the pretraining of embeddings by harnessing pharmaco-kinetic (PK) knowledge, which elucidates the body's reaction to drug absorption. The efficacy of clinical trials is significantly influenced by factors such as the pharmacokinetic properties of a drug and the disease risk profile. In this process, we utilize a spectrum of publicly accessible PK experimental scores. Employing this data, our pretraining is directed toward predictive models for key ADMET (Absorption, Distribution, Metabolism, Excretion, Toxicity) properties. These properties are integral in drug discovery, offering vital insights into the comprehensive interaction of a drug with the human body~\cite{ghosh2016modeling,fu2019pearl}: \ 

\noindent{\bf (1) Absorption} model quantifies the period of a drug's absorption process within the human body.\ 

\noindent{\bf (2) Distribution} model evaluates how efficiently the drug molecules traverse the bloodstream and reach various bodily regions.\

\noindent{\bf (3) Metabolism} model assesses the active duration of the drug's therapeutic effect.\
 
\noindent{\bf (4) Excretion} model gauges the effectiveness of the body in eliminating toxic elements of the drug. \ 
 
\noindent{\bf (5) Toxicity} model appraises the potential adverse effects a drug might have on the human body.
 
For each of these properties, we develop dedicated models to calculate their respective scores and latent embeddings. Our approach involves processing molecular inputs and generating binary outputs, which reflect the presence or absence of the desired ADMET property. 

\begin{equation}
\label{eqn:admet}
\text{\underline{ADMET}}\quad  
\begin{array}{l}
\bfe_* = \Phi_*(\bfe_{\tau}),\ \  \hat{\omega}_* = \sigma(\mathrm{FCNN}(\bfe_{*})) \\
\min\ -{\omega}_{*}\log \hat{\omega}_{*} - (1-{\omega}_{*}) \log (1 - \hat{\omega}_{*}), \\
\end{array}
\end{equation}
where $\bfe_{\tau} \in \RB^d$ is the input drug embedding defined in Eq.~\eqref{eqn:molecule_feature}, ${\omega}_{*}\in\{0,1\}$ is the binary label, $*$ can be A, D, M, E, and T. $\mathrm{FCNN}$ is a one-layer fully connected neural network. $\sigma(\cdot)$ represents the Sigmoid function that maps the output of $\mathrm{FCNN}$ to the binary label ${\omega}_{*}$. $\Phi_{*}$ can be any neural network. Furthermore, we use highway neural network~\cite{srivastava2015highway}, which is denoted as 
\begin{equation}
\label{eqn:highway}
\text{\underline{Highway Network}}\quad  \bfy = \text{highway}(\bfx), \ \ \  \bfy,\bfx\in\RB^d. 
\end{equation}
This choice is motivated by the need to mitigate the vanishing gradient problem, a critical consideration in deep neural network training.

\smallskip
\noindent{\bf Disease risk embedding and trial risk prediction:} Our model extends beyond drug properties, incorporating knowledge gleaned from historical data on trials related to the target diseases. { We integrate information from various sources to assess disease risk: 1) Disease descriptions and their corresponding ontologies, and 2) Empirical data on historical trial success rates for each disease.} We leverage detailed statistics on the success rates of diseases across different phases of clinical trials, as documented by~\cite{Hay2014-qw}, which serve as a supervision signal for training our {\bf trial risk prediction model}. More precisely, we utilize previous trial data,  available at \url{ClinicalTrials.gov}, to predict the likelihood of success for upcoming trials based on the specific disorders involved.
%%% rephrase
% As detailed statistics for the trial success rate of each disease at different trial phases are widely available~\cite{Hay2014-qw}, we will consider that as the supervision signal to train the {\bf trial risk prediction model}. 
% More specifically, given the diseases in the trial, we leverage the historical trial data to predict their success rate, the data is also available at \url{ClinicalTrials.gov}. 
The predicted trial risk, denoted as $\hat{\omega}_{{\Psi}} $, and the embedding,  $\bfe_{\Psi} \in \RB^d$ are derived using a two-layer highway neural network (Eq.~\ref{eqn:highway}) ${\Psi}(\cdot)$:
\begin{equation}
\label{eqn:risk}
\text{\underline{Disease Risk}}
\begin{array}{l}
\bfe_{\Psi} = {\Psi}(\bfe_{\delta}), \ \ \hat{\omega}_{\Psi} = {\sigma}(\mathrm{FCNN}(\bfe_{\Psi})),  \\
\min -{\omega}_{\Psi}\log \hat{\omega}_{\Psi} - (1-{\omega}_{\Psi}) \log (1 -  \hat{\omega}_{\Psi}),    \\
\end{array}
\end{equation} 
where $\bfe_{\delta} \in\RB^d$ is the input disease embedding in Eq.~\eqref{eqn:disease_feature}, $\hat{\omega}_{\Psi}\in[0,1]$ is the predicted trial risk between 0 and 1 (with 0 being the most likely to fail and 1 the most likely to succeed), and ${\omega}_{\Psi} \in \{0,1\}$ is the binary label indicating the success or failure of the trial as a function of disease only. $\mathrm{FCNN}$ is the one-layer fully connected layer. $\sigma(\cdot)$ represents the Sigmoid function that maps the output of $\mathrm{FCNN}$ to the binary label ${\omega}_{\Psi}$.
Binary cross entropy loss between ${\omega}_{\Psi}$ and $\hat{\omega}_{\Psi}$ is used to guide the training. 

\subsection{Hierarchical Interaction Graph}
\label{sec:gnn}

% In this section, we mainly describe (1) the construction of trial interaction graph $\calG$ and (2) how to predict trial approval using \textit{dynamic attentive graph neural network} on $\calG$. 

\noindent\textbf{(I). Trial Interaction Graph}
% \subsubsection{2.5.1 \ Trial Interaction Graph Construction} 
We have devised a {\it hierarchical interaction graph}, denoted as $\mathcal{H}$, which serves as the backbone for establishing connections among all input data sources and the crucial variables that exert influence over the approvals of clinical trials. Below, we provide a comprehensive description of this interaction graph along with its initialization procedure. The interaction graph $\mathcal{H}$ is composed of four distinct tiers of nodes, each of which is intricately interconnected to reflect the intricate development process of real-world clinical trials. These tiers are as follows:
%%% rephrase
% The interaction graph $\calG$ is constructed in a way to reflect the real-world trial development process, and it consists of four tiers of nodes that are connected between tiers: 

\noindent{\bf (1) Input nodes} encompass drugs, target diseases, and trial protocols with node features of input embedding $\bfe_{\tau}$, $\bfe_{\delta}, \bfe_{\pi} \in \RB^d$, indicated in green in Figure~\ref{fig:method} (Section~\ref{sec:input}). \ 

\noindent{\bf (2) External knowledge nodes} include ADMET embeddings $\bfe_{\text{A}}, \bfe_{\text{D}}, \bfe_{\text{M}}, \bfe_{\text{E}}, \bfe_{\text{T}} \in \RB^d$, as well as disease risk embedding $\bfe_{\Psi}$. These representations are initialized with pretrained external knowledge and are indicated in blue in Figure~\ref{fig:method} (Section~\ref{sec:knowledge}). \ 
% pharmaco-kinetics node $\bfh_{PK}$ related to ADMET embeddings, intermediate embeddings for disease $\bfh_d$, molecule $\bfh_m$ and trial protocol $\bfh_p$.

\noindent{\bf (3) Aggregation nodes} include (a) Interaction node $\bfe_{\text{IN}}$ connecting disease $\bfe_{\delta}$, drug molecules $\bfe_{\tau}$ and trial protocols $\bfe_{\pi}$; (b) Pharmaco-kinetics node $\bfe_{\text{PK}}$ connecting ADMET embeddings $\bfe_{\text{A}}, \bfe_{\text{D}}, \bfe_{\text{M}}, \bfe_{\text{E}}, \bfe_{\text{T}}$, $\bfe_T \in \RB^d$ and (c) Augmented interaction node $\bfe_{\text{AU}}$ that augment the interaction node $\bfe_{\text{IN}}$ using disease risk node $\bfe_{\Psi}$. Aggregation nodes are indicated in yellow in Figure~\ref{fig:method}. \

\noindent{\bf (4) Prediction node:} $\bfe_{\text{PR}}$ node serves as the connection point between the Pharmaco-Kinetics node $\bfe_{\text{PK}}$ and the Augmented Interaction node $\bfe_{\text{AU}}$ for making predictions. It is represented in gray in Figure~\ref{fig:method}. The input nodes and external knowledge nodes have been previously detailed, and the resulting representations are utilized as node embeddings within the interaction graph. In the following sections, we elaborate on the aggregation nodes and the prediction nodes.

\smallskip
\noindent{\bf  Aggregation nodes:} The PK (Pharmaco-Kinetics) node aggregates information related to the five ADMET properties (Eq.~\ref{eqn:admet}). We obtain PK (Pharmaco-Kinetics) embedding as follows:
\begin{equation}
\label{eqn:pk}
\begin{aligned}
\text{\underline{PK Embedding}} \\
\bfe_{\text{PK}} &= \mathcal{PK}(\bfe_{\text{A}}, \bfe_{\text{D}}, \bfe_{\text{M}}, \bfe_{\text{E}}, \bfe_{\text{T}}), \\
\bfe_{\text{PK}} &\in \RB^d.
\end{aligned}
\end{equation}
Here, $\mathcal{PK}(\cdot)$ represents a one-layer fully-connected layer ( input dimension is $5*d$, output dimension is $d$ ) , whose input feature concatenating $\bfe_{\text{A}}, \bfe_{\text{D}}, \bfe_{\text{M}}, \bfe_{\text{E}}, \bfe_{\text{T}}$, followed by $d$-dimensional two-layer highway neural network (Eq.~\ref{eqn:highway})~\cite{srivastava2015highway}.

Next, we model the interaction among the input drug molecule, diseases, and protocols through an interaction node, and obtain its embedding as follows:
\begin{equation}
\label{eqn:interaction}
\begin{aligned}
\text{\underline{Interaction Embedding}}\\
\bfe_{\text{IN}} &= \mathcal{IN}(\bfe_{\tau}, \bfe_{\delta}, \bfe_{\pi}), \\
\bfe_{\text{IN}} &\in \RB^{d}, 
\end{aligned}
\end{equation}
where $\bfe_{\tau}, \bfe_{\delta}, \bfe_{\pi}$ represent input embeddings defined in Eq.~\eqref{eqn:molecule_feature}, Eq.~\eqref{eqn:disease_feature} and Eq.~\eqref{eqn:protocol_embedding}, respectively. 
The neural architecture of $\mathcal{IN}(\cdot)$ consists of a one-layer fully-connected network (with input dimension is $3*d$ and output dimension is $d$) followed by a $d$-dimensional two-layer highway network (Eq.~\ref{eqn:highway}) \cite{srivastava2015highway}.

We also employ an augmented interaction model to combine (i) the trial risk associated with the target disease $\bfe_{\Psi}$ (Eq.~\ref{eqn:risk}) and (ii) the interaction among disease, molecule, and protocol represented by $\bfe_{\text{IN}}$ (Eq.~\ref{eqn:interaction}). 
\begin{equation}
\label{eqn:ai}
\text{\underline{Augmented Interaction}}\ \ \ \ \bfe_{\text{AU}} = \mathcal{AU} ({\bfe_{\Psi}, \bfe_{\text{IN}}}), \ \bfe_{\text{AU}} \in \RB^{d}. 
\end{equation}
Here $\mathcal{AU}(\cdot)$ is a one-layer fully connected network (with input dimension is $2*d$, output dimension is $d$) followed by a two-layer $d$-dimensional highway network (Eq.~\ref{eqn:highway}) \cite{srivastava2015highway}.

\smallskip
\noindent{\bf Prediction node} synthesizes the Pharmaco-kinetics and the augmented interaction to derive the final prediction as follows:
\begin{equation}
\label{eqn:trial}
\text{\underline{Trial Prediction}}\ \ \ \ \bfe_{\text{PR}} = \mathcal{PR} (\bfe_{\text{PK}}, \bfe_{\text{AU}} ), \ \bfe_{\text{PR}} \in \RB^d. 
\end{equation}
Similar to $\mathcal{IN}()$ and $\mathcal{AU}()$, the architecture of $\mathcal{PR}$ consists of a one-layer fully connected network (with input dimension is $2*d$, output dimension is $d$) followed by a $d$-dimensional two-layer highway network (Eq.~\ref{eqn:highway}) \cite{srivastava2015highway}.

\smallskip
\noindent\textbf{Training}
After the message-passing phase in the GCN, we obtain updated representations for each trial component. These representations encode the essential information learned from the network. To predict trial success $\hat{\omega}$, we feed the final-layer ($N$-th layer) representation of the trial prediction node into a one-layer fully-connected network with a sigmoid activation function. We employ binary cross-entropy loss for training, which measures the dissimilarity between the predicted values and the true ground truth labels.
\begin{equation}
\label{eqn:bce}
\begin{aligned}
& \hat{\omega} = {\sigma}(\text{FCNN}(\bfe^{N}_{\text{PR}})). \\
& \calL_{\text{classify}} = - {\omega} \log \hat{\omega} - (1-{\omega})\log(1 - \hat{\omega}). 
\end{aligned}
\end{equation} 
In our case, ${\omega} \in \{0,1\}$ represents the ground truth, with ${\omega} = 1$ indicating a successful trial and 0 indicating a failed one.
\mname is trained in an end-to-end manner, optimizing its ability to predict trial approvals based on the learned representations~\cite{fu2022antibody}.
$\sigma(\cdot)$ represents the Sigmoid function.
% For an illustration of the entire framework, we summarize it in Algorithm~\ref{alg:main}. 

\subsection{Selective Classification to Quantify Uncertainty}

\begin{algorithm}[h!]
\caption{\mname Framework with selective classification} 
\begin{algorithmic}[1]
\STATE \# 1. Pretrain 
\STATE Pretrain basic modules: \\
\ \ (i) ADMET models ($\calA, \calD, \calM, \calE, \calT$); \\
\ \ (ii) disease risk (DR) model.  
\STATE Construct Interaction Graph $\mathcal{H}$. 
\STATE \# 2. Train \mname on $\mathcal{X}_{train}$
% \STATE Build \mname as Figure~\ref{fig:method}. 
\STATE minimize $\calL_{\text{classify}}$ (Eq.~\ref{eqn:bce}), update the remaining part of model. 
\STATE \# 3. Find threshold
\STATE Specify $\alpha$, $\beta$, and the calibration set. Pick the threshold $\hat{\lambda}$ over the calibration set by the empirical selective accuracy (Eq.~\ref{eqn:selective risk}).

\STATE \# 4. Inference 
\STATE Given new data $(\mathcal{T}, \mathcal{D}, \mathcal{P})$, predict success probability $\hat{\omega}$.
\STATE \# 5. Classify selectively
\STATE Output the prediction if $\max(\hat{\omega}, 1-\hat{\omega})$ exceeds $\hat{\lambda}$, otherwise abstains.
\end{algorithmic}
\label{alg:main2}
\end{algorithm}

\begin{figure*}[h!]
\centering
\includegraphics[width=\textwidth]{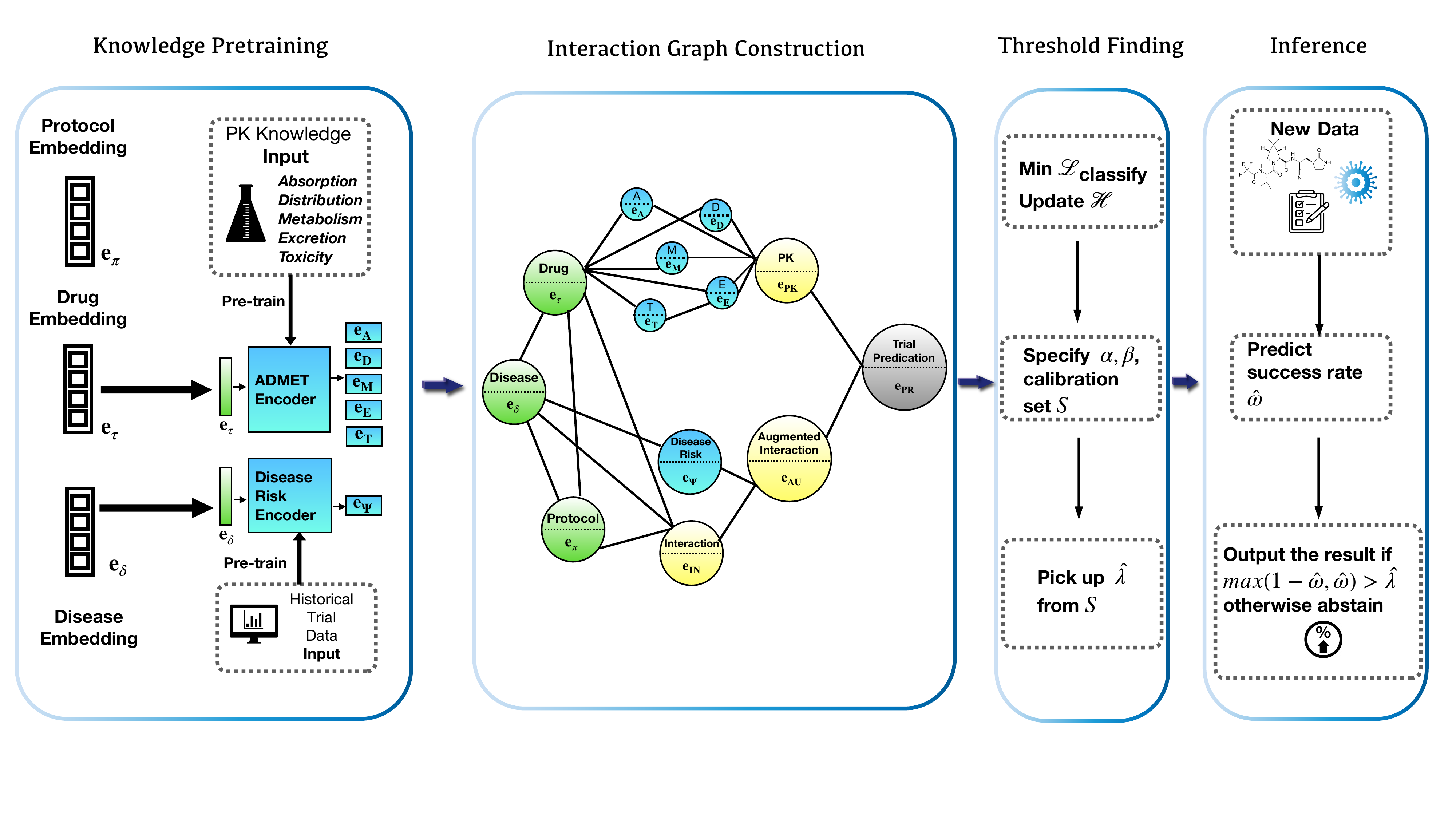}
\caption{Selective Classification on HINT. }
\label{fig:selective_classification}
\end{figure*}

% We consider a binary classification problem, such as clinical trial outcome prediction, which HINT was designed to solve. Let $\mathcal{X}$ be some feature space (e.g., trial embeddings) and $\mathcal{Y}$ be the label set, $\mathcal{Y} = \{0, 1\}$. Let $P(X, Y)$ be a joint distribution over $\mathcal{X} \times \mathcal{Y}$. The selective classifier $(f,g)$ is made up of the selective function $g: \mathcal{X} \rightarrow \{0,1\}$ and the classifier $f$, which produces a probability for each label provided in the input $x$.

% \[\left.(f,g)(x)\triangleq \left\{\begin{array}{ll}f(x),&\text{if}g(x)=1;\\\text{abstain},&\text{if}g(x)=0.\end{array}\right.\right.\]

We consider a binary classification problem, such as clinical trial outcome prediction, which HINT was designed to solve. Let $f: \mathcal{X} \rightarrow \mathcal{Y}$ be the model, with feature space $\mathcal{X}$ (e.g., trial embeddings) and label set $\mathcal{Y} = \{0, 1\}$. Let $P(X, Y)$ be a joint distribution over $\mathcal{X} \times \mathcal{Y}$. The selective classifier $(f,g)$ is made up of the selective function $g: \mathcal{X} \rightarrow \{0,1\}$ and the classifier $f$, which produces a probability for each label provided in the input $x$.

\[\left.(f,g)(x)\triangleq \left\{\begin{array}{ll}f(x),&g(x)=1;\\\text{abstain},&g(x)=0.\end{array}\right.\right.\]

Therefore, the prediction of input $x$ is abstained if $g(x) = 0$. We measure the performance of a selective classifier using \textit{coverage} and \textit{risk}. Coverage is defined as the probability mass of the non-rejected region of $\mathcal{X}$, $\psi(f,g)=\mathbb{E}[g(X)]$. Given a loss function $\mathcal{L}$, the selective risk of ($f, g$) is defined as 
\[
\mathcal{R}(f,g)=\frac{\mathbb{E}[\mathcal{L}(f(x),y)g(x)]}{\psi(f,g)}.
\]

\paragraph{Selective Classification (SC)}
In many scenarios, it is preferable to display a model's predictions only when it has high confidence. For instance, in medical diagnosis, we might only want the model to make predictions if it is 90\% certain, and if not, it should say, ``I'm uncertain'' The algorithm demonstrated below strategically abstains in order to achieve higher accuracy in clinical trial outcome prediction tasks. 

More formally, given sample-label pairs $\{(X_i,Y_i)\}_{i=1}^n$ and a clinical trial outcome predictor $\hat{f}$ , we seek to ensure 
\begin{equation}
\label{eqn:selective accuracy guarantee}
    \mathrm{P}(Y_{test}=\widehat{Y}(X_{test})\mid\widehat{P}(X_{test})\geq\hat{\lambda})\geq1-\alpha, 
\end{equation} 
where $\widehat{Y}(x)=\arg\max_{y}\hat{f}(x)_{y}$, $\widehat{P}(X_{test})=\max_{y}\hat{f}(x)_{y}$, and $\hat{\lambda}$ is a threshold calculated upon the calibration set. The accuracy computed over only a subset of high-confidence predictions is called a \emph{selective accuracy} guarantee.
We pick the threshold based on the empirical estimate of selective accuracy on the calibration set.

% \[\widehat{R}(\lambda)=\frac{1}{n(\lambda)}\sum_{i=1}^n\mathbbm{1}\left\{Y_i\neq\widehat{Y}(X_i)\text{and}\widehat{P}(X_i)\ge\lambda\right\},\text{where } n(\lambda)=\sum_{i=1}^n\mathbbm{1}\left\{\widehat{P}(X_i)\ge\lambda\right\}.\]

% \usepackage{bbm}

\begin{equation}
\label{eqn:selective risk}
\begin{split}
&\widehat{R}(\lambda) = \frac{1}{n(\lambda)}
 \sum_{i=1}^n{1} \left\{Y_i\neq\widehat{Y}(X_i)\text{ and }\widehat{P}(X_i)\ge\lambda\right\}, \\
&\text{where } n(\lambda) = \sum_{i=1}^n {1} \left\{\widehat{P}(X_i)\ge\lambda \right\}.
\end{split}
\end{equation}
where ${1}(\cdot)$ is an indicator function. 

In particular, we will scan across values of $\lambda$, looking at a conservative upper bound for the true risk (i.e., the top end of a confidence interval for the selective misclassification rate). Realizing that $\widehat{R}(\lambda) $ is a Binomial random variable with $n(\lambda)$ trials, we upper-bound the misclassification error as

\begin{equation}
\label{eqn:binomcdf}
    \widehat{R}^+(\lambda)=\sup\left\{r:\text{BinomCDF}(\widehat{R}(\lambda);n(\lambda),r)\geq\beta\right\}
\end{equation}

for some user-specified failure rate $\beta\in[0, 1]$. Then, scan the upper bound until the last time the bound exceeds $\alpha$,
\begin{equation}
\label{eqn:lambdahat}
\hat{\lambda}=\inf\left\{\lambda:\widehat{R}^+(\lambda')\leq\alpha\text{ for all }\lambda'\geq\lambda\right\}.
\end{equation}

Using $\hat{\lambda}$ will satisfy Equation~(\ref{eqn:selective accuracy guarantee}) with high probability.

\subsection{{Dataset}}
\label{sec:dataset}
In our study, we utilized the clinical trial outcome prediction benchmark dataset, encompassing three phases.
We employed the TOP clinical trial outcome prediction benchmark presented by~\cite{fu2022hint,fu2023automated}. This dataset encompasses information on drugs, diseases, trial protocol, and trial outcomes for a total of 17,538 clinical experiments. These trials are categorized into three phases: Phase I with 1,787 trials, Phase II with 6,102 trials, and Phase III with 4,576 trials. Success rate differs across phases: 56.3\% in Phase I, 49.8\% in Phase II, and 67.8\% in Phase III. A breakdown of the diseases targeted can be found in Table 1. Our research is executed distinctly in each phase of the trials.

\begin{figure}[h!]
\centering
\includegraphics[width=\columnwidth]{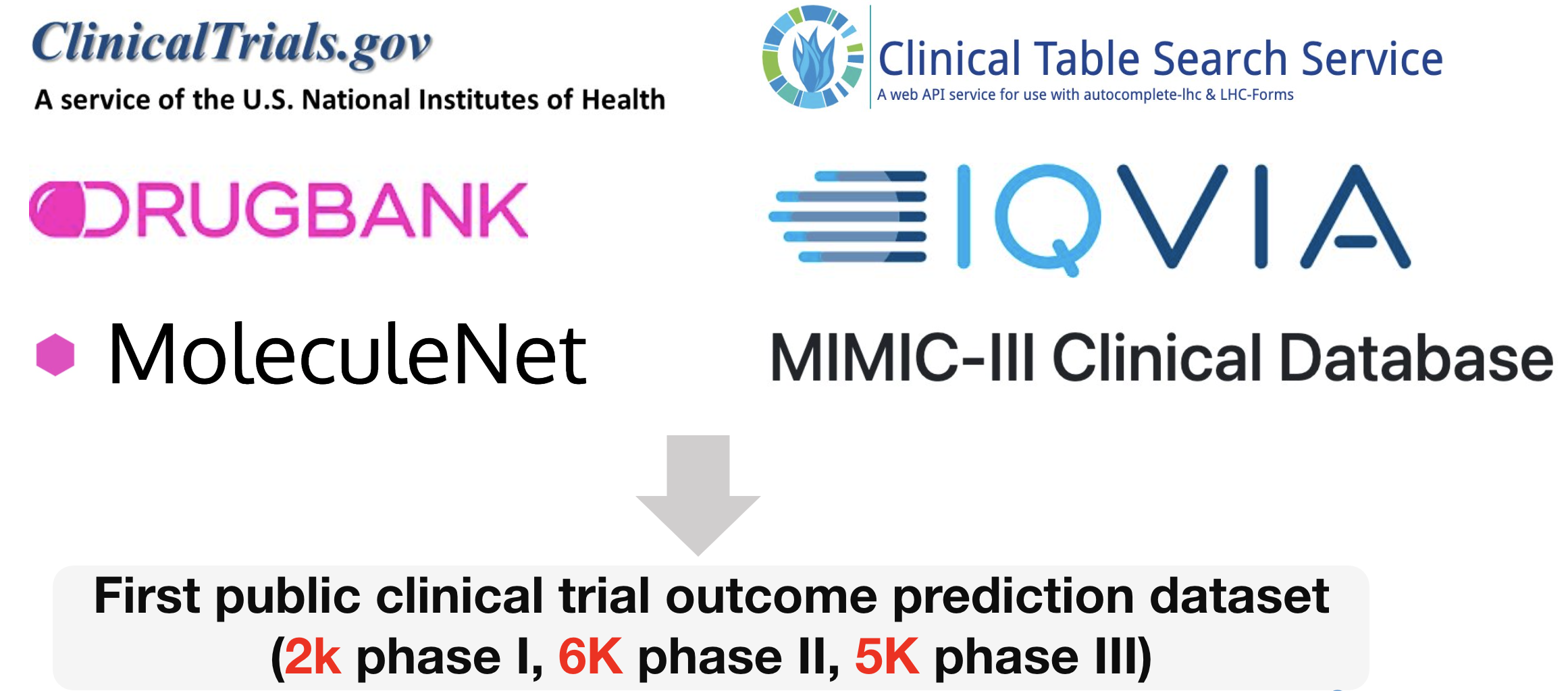}
\caption{Dataset. The dataset is curated by aggregating multi-modal data {from various sources}. This dataset contains data on medications ({drug molecules}), diseases, trial protocols ({text data}), {and approval information (labels)}. }
\label{fig:data}
\end{figure}

\noindent\textbf{Data Processing and Linking.} Next, we describe how to process and link the parsed trial data to machine learning-ready input and output format: 
% \begin{itemize}
\\\noindent$\bullet$ {\it Drug molecule data} are extracted from \url{https://clinicaltrials.gov/} and linked to its molecule structure (SMILES strings and the molecular graph structures) using DrugBank Database~\cite{wishart2018drugbank} (\url{https://www.drugbank.com/}). 
\\\noindent$\bullet$ {\it Disease data} are extracted from \url{https://clinicaltrials.gov/} and linked to ICD-10 codes and disease description using \url{clinicaltables.nlm.nih.gov}  and then to CCS codes via \url{hcup-us.ahrq.gov/toolssoftware/ccs10/ccs10.jsp}. 
\\\noindent$\bullet$  {\it Trial protocol data} are extracted from \url{https://clinicaltrials.gov/}, in particular, the study description section, outcome section, and eligibility criteria section. 
\\\noindent$\bullet$  {\it Trial approval data} (binary label) are available at TrialTrove (\url{https://pharmaintelligence.informa.com/products-and-services/data-and-analysis/trialtrove}).
\\\noindent$\bullet$ {\it Auxiliary drug pharmacokinetics data} include five datasets across the main categories of PK. For absorption, we use the bioavailability dataset provided in Ma et al. paper supplementary~\cite{bioavail}. For distribution, we use the blood-brain-barrier experimental results provided in Adenot et al. study~\cite{BBB}. For metabolism, we use the CYP2C19 experiment from Veith et al.~\cite{CYP} paper, which is hosted in the PubChem bioassay portal under AID 1851. For excretion, we use the clearance dataset from the eDrug3D database~\cite{clearance}. For toxicity, we use the ToxCast dataset~\cite{richard2016toxcast}, provided by MoleculeNet (\url{http://moleculenet.ai}). We consider drugs that are not toxic across all toxicology assays as not toxic and otherwise toxic.
   % \item {\it Auxiliary disease risk data}
% \end{itemize}
Concretely, we collected all the clinical trial data from \url{https://clinicaltrials.gov/}. The historical trial approval probability on each disease ({\it disease risk} in our model) is also extracted from this data source. 
{For drug knowledge, the data are extracted from multiple public sources. We obtain drugs' molecule information from DrugBank Database~\cite{wishart2018drugbank} (\url{https://www.drugbank.com/}).}
For drug property knowledge, 
(3) We obtain diseases' ICD-10 code from \url{clinicaltables.nlm.nih.gov}. \\

{
\paragraph{Temporal data split based on start and completion date.}
We leverage temporal split, which refers to splitting the data samples based on their time stamps. The earlier data samples are used for training and validation, while the later data are used for testing. 
The later trials would leverage the knowledge or results obtained in the earlier trials. 
To make sure that there is no \textit{information leakage} and fit the real clinical trial outcome prediction setup, we leverage temporal split when partitioning the whole data into training/validation/testing datasets, following~\cite{fu2022hint,fu2023automated}. 
Specifically, as mentioned, all the trials we use in the learning process have both a start date and a completion date. We set up a split date to ensure that all the trials in the training and validation dataset are complete before the split date and all the trials in the test set are started after the split date~\cite{du2023abds}. 
For instance, in Phase I, we trained the model on trials before Aug 13th, 2014, and tested on trials post this date, as shown in Table~\ref{table:statistics}. }

\begin{table}[h!]
\centering
\caption{Data statistics of clinical trial outcome prediction benchmark dataset. During training, we randomly select 15\% training samples for model validation. {The earlier trials are used for learning, while the later trials are used for inference.} ``Succ'', and ``Fail'' are abbreviations for ``success'' and ``failure'', respectively. }
\label{table:statistics}
% \resizebox{\columnwidth}{!}{
\begin{tabular}{l|ccccc}
\toprule
\multirow{2}{*}{Settings}  & \multicolumn{2}{c}{Train} & \multicolumn{2}{c}{Test} & \multirow{2}{*}{{Split Date}} \\
% \cline{2-5}
& Succ & Fail & Suss & Fail &  \\ \midrule 
Phase I  & 702 & 386 & 199 & 113  &  08/13/2014 \\
Phase II  & 956 & 1655 & 302 & 487 & 03/20/2014 \\
Phase III  & 1,820 & 2,493 & 457 & 684 & 04/07/2014 \\
\bottomrule
\end{tabular}
\end{table}

\begin{figure*}[h!]
\centering
\includegraphics[width=\textwidth]{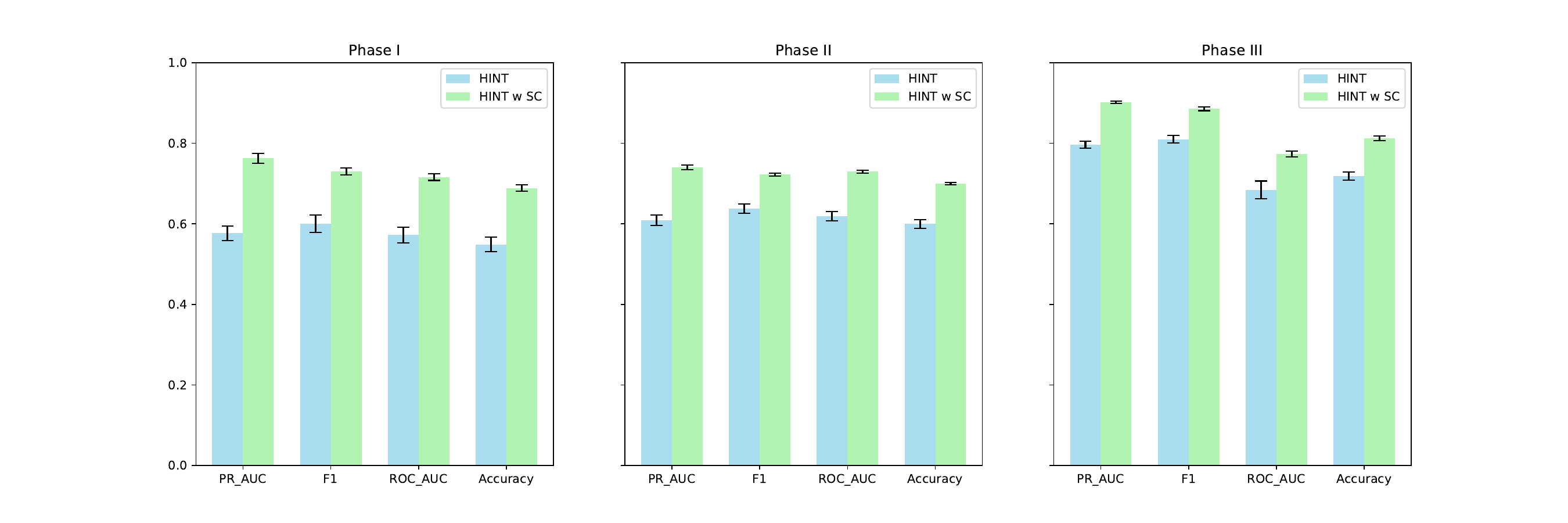}
\caption{Phase-Level outcome prediction (Average accuracy and standard deviation). Our method significantly outperforms HINT (i.e., passing hypothesis testing, the p-value is smaller than 0.05) in all the metrics in all the tasks. }
\label{fig:Phase-Level_Outcome}
\end{figure*}

\begin{figure*}[h!]
\centering
\includegraphics[width=\textwidth]{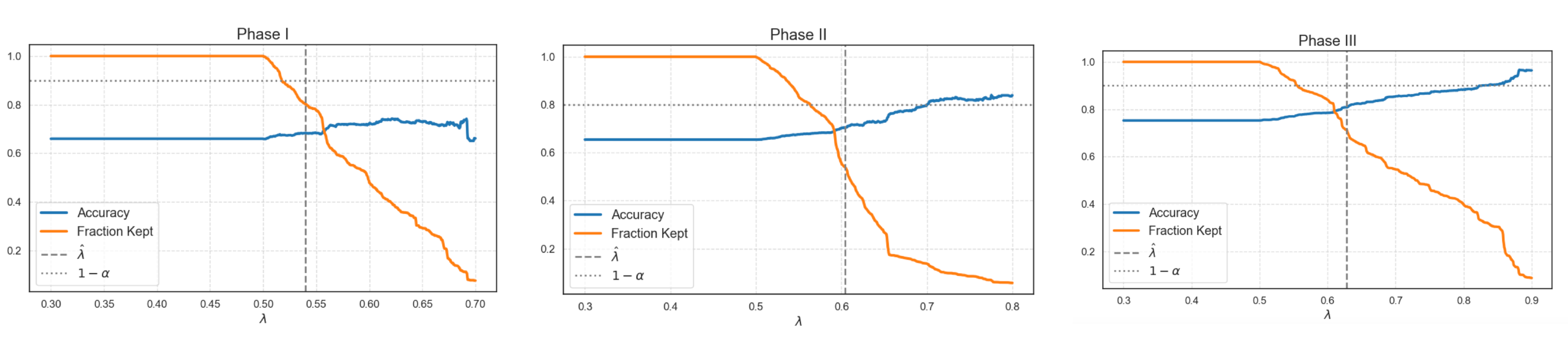}
\caption{Tradeoff between Selective Accuracy and Fraction Kept. }
\label{fig:selective_accuracy}
\end{figure*}

\subsection{{Evaluation}}

Our evaluation utilized various metrics, such as PR-AUC, F1, ROC-AUC, and Accuracy. PR-AUC assesses the model's ability to differentiate between positive and negative examples. 

\noindent\textbf{PR-AUC}: (Area Under the Precision-Recall Curve). It quantifies the area under the precision-recall curve, representing how well the model separates positive and negative examples. PR-AUC focuses on the trade-off between precision (positive predictive value) and recall (true positive rate) across different probability thresholds.

\noindent\textbf{F1}: The F1 score represents the harmonic mean of precision and recall. The F1 score is a single metric combining precision and recall into a single value to assess a classification model's performance. The F1 score adeptly balances the trade-off between precision (the accuracy of positive predictions) and recall (the ability to identify all positive instances), ensuring a more accurate evaluation of the model's accuracy.

\noindent\textbf{ROC-AUC}: (Area Under the Receiver Operating Characteristic Curve) It focuses on the trade-off between true positive rate (TPR or recall) and false positive rate (FPR) across different probability thresholds. ROC-AUC quantifies the area under the ROC curve, which is a plot of TPR against FPR~\cite{lu2022cot}. A higher ROC-AUC value indicates better model discrimination and the ability to distinguish between positive and negative examples.

\noindent\textbf{Accuracy}: The ratio of correct predictions to the total number of samples.

We report the results of hypothesis testing in terms of p-value to showcase the statistical significance of our method over the best baseline results. If the p-value is smaller than 0.05, we claim our method significantly outperforms HINT. 

Furthermore, to promote transparency and facilitate reproducibility in the scientific community, we have made our code publicly available at the provided GitHub link\footnote{\url{https://github.com/Vincent-1125/Uncertainty-Quantification-on-Clinical-Trial-Outcome-Prediction}}.

\section{{Results}}

\subsection{Quantitative Results}

{We only include HINT as the baseline model. The reason is that HINT outperforms a bunch of baseline methods (including traditional machine learning methods like logistic regression, random forest, AdaBoost and deep learning methods such as DeepEnroll, COMPOSE, etc) across various phases statistically significantly~\cite{fu2023automated}. }

We conduct experiments to evaluate the effect of using selective classification (SC) on trial outcome prediction in the following aspects: 

% Phase I
\begin{table}[]
\caption{Phase-Level outcome prediction. ``*'' means our method significantly outperforms HINT (i.e., passing hypothesis testing, the p-value is smaller than 0.05.) in all the metrics in all the tasks. }
\small
\begin{tabular}{lccc}
\toprule
\multicolumn{4}{c}{Phase I}                               \\
\midrule
            & HINT          & HINT with SC  & Improvement \\
PR-AUC      & 0.5765±0.0119 & 0.7631±0.0119* & 32.37\%    \\
F1          & 0.6003±0.0091 & 0.7302±0.0091* & 21.64\%    \\
ROC-AUC     & 0.5723±0.0084 & 0.7164±0.0084* & 25.18\%    \\
Accuracy    & 0.5486±0.0046 & 0.6885±0.0083* & 25.50\%    \\
Retain rate & /             & 0.7874±0.0267 & /          \\
\end{tabular}
% \end{table}

% Phase II
% \begin{table}[]
\small
\begin{tabular}{lccc}
\midrule
\multicolumn{4}{c}{Phase II}                              \\
\midrule
            & HINT          & HINT with SC  & Improvement \\
PR-AUC      & 0.6093±0.0131 & 0.7399±0.0055* & 21.43\%    \\
F1          & 0.6377±0.0110 & 0.7224±0.0036* & 13.28\%    \\
ROC-AUC     & 0.6191±0.0116 & 0.7299±0.0038* & 17.90\%    \\
Accuracy    & 0.5998±0.0052 & 0.7002±0.0031* & 16.74\%    \\
Retain rate & /             & 0.5414±0.0021 & /          \\
\end{tabular}
% \end{table}

% Phase III
% \begin{table}[]
\small
\begin{tabular}{lccc}
\midrule
\multicolumn{4}{c}{Phase III}                             \\
\midrule
            & HINT          & HINT with SC  & Improvement \\
PR-AUC      & 0.7965±0.0092 & 0.9022±0.0031* & 13.27\%    \\
F1          & 0.8098±0.0093 & 0.8857±0.0048* & 9.37\%     \\
ROC-AUC     & 0.6843±0.0220 & 0.7735±0.0077* & 13.04\%    \\
Accuracy    & 0.7190±0.0063 & 0.8122±0.0059* & 18.69\%    \\
Retain rate & /             & 0.7117±0.0172 & /          \\
\bottomrule
\end{tabular}
\label{tab:result}
\end{table}

% To examine whether selective classification has an obvious improvement over the original model, we first conducted phase-level outcome prediction. For each phase, a single model was trained to make the prediction. Considering the experiments' reproducibility and the results' comparability, instead of retraining three separate models on phase I, II, and III, we used the pre-trained models from the HINT repository. We implemented selective prediction according to the previously mentioned methodology. Simply put, we select a certain $\lambda$ over the calibration set, which is the training set mentioned above in this context, and then we set the threshold to determine whether we retain or abstain a certain prediction made by the model. We would retain a prediction if the softmax output of a given sample exceeds this threshold and, otherwise, abstain it. 

We conducted a phase-level outcome prediction to discern the enhancement offered by selective classification over the conventional model. Each trial phase was modeled individually using pre-trained models from the HINT repository to ensure consistent and reproducible outcomes. {We incorporated selective classification by setting a calibrated threshold $\lambda$ on the training set. This threshold acted as a decision boundary to either retain or abstain from predictions based on the model's confidence, as indicated by the softmax output. } 

The detailed results and observations are presented in Table~\ref{tab:result}. The results reveal that:
(1) We observed significant improvements across all phases, with Phase I showing the most notable improvements. This indicates a strong adaptability of our model to early-stage trials. 
(2) All key performance metrics demonstrated marked improvements. The most striking gains are observed in PR-AUC. Although the F1 score's enhancements were comparatively modest, they are indicative of a meaningful improvement in the model's ability to maintain a balance between precision and recall—a critical consideration in the realm of imbalanced clinical trial datasets.

The results indicate a consistent enhancement through the phases with selective classification (SC). Phase I trials show a remarkable 32.37\% increase in PR-AUC, indicating a substantial boost in the model’s precision and recall trade-off. Phase II and III also show notable improvements, albeit less pronounced than Phase I. This could be due to the higher initial success rates in later phases, which leave less room for improvement. The data suggests that SC has the most significant impact where the uncertainty in predictions is greatest, thereby emphasizing the utility of SC in early-stage trials where risk assessment is critical.

We also tune $\lambda$ and show the change of selective accuracy and fraction kept (coverage) in Figure~\ref{fig:selective_accuracy}. We find by increasing $\lambda$, accuracy would grow, and the fraction kept would decrease. That is to say, there is a tradeoff between selective accuracy and fraction kept, as expected. We need to select appropriate $\lambda$ to get a balance between them.

To show our method can predict clinical trial approval accurately and potentially save huge unnecessary costs in the case of failure. 
In 2019, Entresto emerged as a highly anticipated medication for heart failure, the principal cause of mortality in the United States. Backed by Novartis, Entresto was projected to reach peak sales of 5 billion dollars. Despite this, a comprehensive Phase III trial conducted across multiple countries, involving 4,822 patients, yielded disappointing results. The drug failed to decrease mortality rates or achieve any other intended outcomes. Spanning from 2014 to 2019, this five-year trial incurred an estimated cost of 200 million dollars\footnote{We estimate the cost by multiplying the median cost per patient by the total number of patients~\cite{moore2018estimated}. }. 
Next, we evaluate whether our approach can foresee such a failure in advance. By inputting the drug (Entresto), the condition (heart failure), and the eligibility criteria into our system, it forecasts a low probability of approval, at just 0.287. This suggests that our method might be capable of providing early warnings to healthcare professionals about the probable lack of success. 

% of the trial and save all these times and costs. 
% We also tested our method on Fevipiprant, which was expected to be Novartis's blockbuster for asthma. The Phase III trial of Fevipiprant took four years (2015-2019) and enrolled 894 patients, which also incurred huge costs (estimated 40 million dollars). Unfortunately, the primary endpoint was not met and Fevipiprant was retired. We feed the drug (Fevipiprant), disease (asthma), and eligibility criteria into our method, and it predicts a 0.380 approval probability, which is low. 

% Similarly, for a recent Phase II study on the effect of Pembrolizumab and Epacadostat on Non-Small Cell Lung Cancer by Incyte and Merck, our method correctly predicts the failure of the trial, among many other examples. 
 
% Our method can also predict the approval probability of the trial accurately, reassuring the drug developers for the prospect of the treatment. For example, our method predicts several recent huge trial successes: Sitagliptin for diabetes by Merck, Etanercept for Rheumatoid Arthritis by Amgen, Afibercept for Glaucoma by Bayer. Notably, we also see that our method can predict accurately in a variety of disease groups, which corroborates our claim that our method is a general clinical trial approval probability prediction model. 

\subsection{Results for Disease Groups}

Our method's effectiveness is assessed across various disease categories, such as cancer/neoplasm/tumor, chronic disease, pain, and cardiovascular disease, with the findings detailed in Table~\ref{table:disease_subgroup}. It is noted that forecasts for cancer/neoplasm/tumor approvals are particularly challenging, exhibiting notably lower accuracy compared to other groups. In contrast, predictions for cardiovascular disease trials show the highest accuracy rate among all categories~\cite{chen2021data}. Trials related to pain and chronic diseases also demonstrate strong predictive performance.

\begin{table}[h!]
\centering
\caption{Results on different disease groups. }
\label{table:disease_subgroup}
\resizebox{\columnwidth}{!}{
\begin{tabular}{lcccc}
\toprule
Cohorts & \% in test set & PR-AUC & F1 & ROC-AUC \\
\midrule
Neoplasm & 13\%  & 0.58{\stdfontsize$\pm$0.01} & 0.56{\stdfontsize$\pm$0.01} & 0.65{\stdfontsize$\pm$0.02} \\ 
Respiratory & 9\% & 0.85{\stdfontsize$\pm$0.02} & 0.87{\stdfontsize$\pm$0.02} & 0.83{\stdfontsize$\pm$0.01} \\
Digestive & 9\% &  0.80{\stdfontsize$\pm$0.01}  & 0.81{\stdfontsize$\pm$0.01} & 0.87{\stdfontsize$\pm$0.00} \\
Nervous system & 11\% & 0.68{\stdfontsize$\pm$0.01}  &  0.79{\stdfontsize$\pm$0.01}  & 0.79{\stdfontsize$\pm$0.01} \\ 
\bottomrule
\end{tabular}}
\end{table}

\section{{Discussion}}

\subsection{{Related Work}}

{\textbf{Clinical Trial outcome prediction.}
Publicly accessible data sources offer crucial insights for forecasting clinical trial approvals. The ClinicalTrials.gov database (publicly available at \href{https://clinicaltrials.gov/}{https://clinicaltrials.gov/}), for example, lists 369,700 historical trials with significant details about them. Furthermore, the standard medical codes for diseases and their descriptions are available through the National Institutes of Health website(publicly available at \href{https://clinicaltables.nlm.nih.gov/}{https://clinicaltables.nlm.nih.gov/}). The DrugBank database (publicly available at \href{ https://www.drugbank.ca/}{ https://www.drugbank.ca/}) provides biochemical profiles of numerous drugs, aiding in the computational modeling of these compounds. }

{In recent years, there have been various preliminary attempts to predict specific aspects of clinical trials, aiming to enhance prediction. These include using electroencephalographic (EEG) measurements to gauge the impact of antidepressant therapies on alleviating depression symptoms~\cite{raj20evaluation}, enhancing drug toxicity predictions through drug and target property characteristics~\cite{hong20predicting,yi2018enhance}, and leveraging phase II trial findings to forecast phase III trials results~\cite{pmlr-v106-qi19a}. Recently, there's a growing inclination towards creating a universal strategy for predicting clinical trial approvals. As a preliminary effort, \cite{Lo2019Machine} ventured beyond refining singular components, opting instead to forecast trial results for 15 ailment categories solely based on disease attributes through statistical analysis.} \cite{wang2024twin} designs digital twins to mimic clinical trials and predict the outcome. 

{Notably, the work of~\cite{fu2022hint,fu2023automated} stands out in this field. Their contributions are three-fold:  
\begin{enumerate}
 \item They established a formal modeling framework for clinical trial outcome prediction, integrating information on drugs, diseases, and trial protocols. 
 \item By utilizing a comprehensive dataset from various online sources, including drug repositories, standardized disease codes, and clinical trial records, they have established a publicly available dataset TOP, based on which researchers can conduct general clinical trial outcome prediction. 
 \item They developed HINT (Hierarchical Interaction Network for Clinical Trial outcome prediction), a machine learning approach that explicitly models the components of clinical trials and constructs the intricate relationships among them. This method surpasses a range of traditional machine learning and deep learning models in performance.
\end{enumerate}}

{
\noindent\textbf{Uncertainty Quantification.}
Regarding the prediction of clinical trial approvals, uncertainty quantification plays a pivotal role, as it aids in assessing the likelihood of trial success and informs decision-making processes. In this context, one principled framework that stands out is conformal prediction~\cite{vovk2005line,papadopoulos2002inductive,lu2023machine}, a versatile and straightforward approach for generating prediction sets applicable to any model. Additionally, selective classification, particularly suitable in binary classification scenarios, opts for abstaining from predictions when confidence is lacking. The idea of abstaining when the model is not certain originated in the last century~\cite{chow1957optimum,hellman1970nearest}. More approaches were proposed in recent years, including using softmax probabilities~\cite{geifman2017selective}, using dropout~\cite{gal2016dropout}, and using deep ensembles~\cite{lakshminarayanan2017simple}. Others incorporated abstention into model training~\cite{bartlett2008classification,geifman2019selectivenet,feng2019selective} and learned to abstain on examples human experts were more likely to get correct~\cite{raghu2019direct,mozannar2020consistent,de2020regression}. On the theoretical level, early work characterized optimal abstention rules given well-specified models~\cite{chow1970optimum,hellman1970probability}, with more recent work on learning with perfect precision~\cite{el2010foundations,khani2016unanimous} and guaranteed risk~\cite{geifman2017selective}.}

\subsection{{Discussion of Results}}
{The quantitative results reveal a consistent improvement across the stages with the application of selective classification (SC). Specifically, Phase I trials exhibit an impressive 32.37\% uplift in PR-AUC, highlighting a significant enhancement in the model's precision and recall balance. While Phases II and III also demonstrate discernible advancements, these are not as marked as in Phase I. This discrepancy may stem from the higher base success rates in the later stages, which naturally offer narrower margins for enhancement. The evidence points to SC having the most pronounced effect in scenarios with the highest prediction uncertainty, underlining SC's value in early-phase trials where accurate risk evaluation is paramount. }

\subsection{{Contributions}}

{The major contributions of this paper can be summarized as:
\begin{itemize}
\item {Methodology: This paper introduces a novel approach that combines selective classification with the Hierarchical Interaction Network (HINT), enhancing the model’s ability to withhold predictions in uncertain scenarios. Our model is built upon fundamental, trustworthy modules and employs a transparent modeling process throughout its formulation.}
\item {Experimental results: Through comprehensive experiments, the paper demonstrates that this approach significantly improves performance metrics. Specifically, the proposed method achieved 32.37\%, 21.43\%, and 13.27\% relative improvement in PR-AUC over the base model (HINT) in phase I, II, and III trial outcome prediction, respectively. When predicting phase III, our method reaches 0.902 PR-AUC scores. }
\item{Applications: The methodology presented has a specific focus on clinical trial outcome predictions, highlighting its potential impact in this critical area of medical research.}
\end{itemize}}

\section{Conclusion}

In conclusion, our study utilizing the Hierarchical Interaction Network (HINT) has presented a transformative approach in the domain of clinical trial outcome prediction. By integrating the selective classification methodology for clinical trial outcome predictions, we have addressed and quantified the inherent model uncertainty, which has illustrated marked enhancements in performance. 

The empirical results are compelling, demonstrating that selective classification confers a significant advantage, particularly evidenced by the pronounced improvements in PR-AUC across all phases of clinical trials. This is indicative of a more discerning model, capable of delivering higher precision in its predictions, especially in the critical early phases of clinical development.

The selective classification's impact is most striking in Phase I trials, where the model's adaptability is crucial due to the higher uncertainty and variability. Despite the smaller gains in the F1 score, the consistent uplift across all metrics, including ROC-AUC and accuracy, underscores the overall increase in the model's predictive reliability.

{\paragraph{Data Availability} 
All the data are publicly available\footnote{\url{https://github.com/futianfan/clinical-trial-outcome-prediction}}. 
The code is publicly available\footnote{\url{https://github.com/Vincent-1125/Uncertainty-Quantification-on-Clinical-Trial-Outcome-Prediction}}. }

{\paragraph{Funding} Work related to this study is supported by the Sontag Foundation (C.V.R), the American Cancer Society (C.V.R.) and the Department of Defense Breast Cancer Research Program (C.V.R.). }

% \clearpage
% \tiny
% \scriptsize   
% \footnotesize
\bibliographystyle{named}
\bibliography{ref}

\end{document}